\documentclass[letterpaper]{article}
\PassOptionsToPackage{table}{xcolor}
\usepackage[preprint]{aaai2027}
\usepackage[hyphens]{url}
\usepackage{graphicx}
\usepackage{natbib}
\usepackage{caption}
\usepackage{microtype}
\usepackage{amsmath}
\usepackage{amssymb}
\usepackage{booktabs}
\usepackage{multirow}
\usepackage{array}
\usepackage{pifont}

\newcommand{\cmark}{\ding{51}}
\newcommand{\xmark}{\ding{55}}
\newcommand{\ind}{\mathbb{I}}
\newcommand{\match}{\operatorname{Match}}
\newcommand{\clip}{\operatorname{clip}}

\definecolor{HeaderGray}{RGB}{224,224,224}
\definecolor{HighlightGray}{RGB}{236,236,236}
\definecolor{LightGray}{RGB}{247,247,247}
\definecolor{RuleGray}{RGB}{90,90,90}

\title{StructReward: Efficient Structured Process Rewards for Self-Correcting Multimodal Reasoning}
\author{Yifan Li\textsuperscript{*} \quad Ruxin Sun\textsuperscript{*} \quad Tongzhou Zhao}
\affiliations{}

\ifPDFTeX
\else\ifXeTeX

\fi
\fi

\begin{document}

\maketitle
\begingroup
\renewcommand{\thefootnote}{*}
\footnotetext{These authors contributed equally.}
\endgroup

\begin{abstract}
\normalfont\normalsize
Reinforcement learning with verifiable rewards (RLVR) has emerged as an effective approach for improving multimodal reasoning. However, most existing methods evaluate an entire response using a binary reward based only on final-answer correctness, thereby discarding the supervision available in intermediate reasoning steps. Process reward models offer finer-grained feedback, but they typically rely on separately trained verifiers, costly chain-of-thought annotations, or online judging by large language models (LLMs). In this work, we introduce \textbf{StructReward}, a compute-efficient framework that provides dense reinforcement signals through structured step-level reward alignment. StructReward represents each generated solution as a sequence of reasoning steps and aligns them with process-labeled reference steps using lightweight numerical, symbolic, and lexical matching rules. The aligned labels are aggregated into a dense process reward and combined with final-answer consistency and output-validity rewards through a gated Group Relative Policy Optimization (GRPO) objective. We further recycle policy rollouts into complementary supervision for response comparison and reflective self-correction, rather than discarding them after policy updates. Separately, we use a strong LLM to rewrite sampled correct trajectories into reflection-oriented training instances, further strengthening the policy's ability to evaluate and refine its reasoning. Since reward computation is performed online without an additional learned verifier or external LLM judge, StructReward substantially reduces the computational overhead of multimodal reinforcement learning. Across the reported model--benchmark comparisons, StructReward improves accuracy by an average of 4.0 percentage points over the corresponding Qwen3-VL-Instruct baselines. Experimental results show that structured process supervision and rollout recycling provide an efficient path toward self-improving multimodal reasoning.
\end{abstract}

\section{Introduction}

Recent vision-language models (VLMs) combine high-resolution visual perception with long-form language reasoning and achieve strong performance on general and expert-level multimodal benchmarks \cite{wang2025mcotsurvey,zhang2026posttraining,bai2025qwen3vl}. Beyond supervised fine-tuning, RLVR has emerged as a scalable paradigm for mathematical and multimodal reasoning. Its automatically checkable task rewards reduce dependence on human preference annotations, while GRPO estimates advantages from groups of responses without a separately learned value model \cite{shao2024deepseekmath,deepseekai2025r1}.

Most multimodal RLVR methods assign a binary outcome reward to an entire response according to final-answer correctness. Although inexpensive, this signal cannot distinguish answer-correct responses with reliable intermediate results from responses that arrive at the same answer through an inconsistent or accidental process. Process reward models (PRMs) provide finer-grained feedback by evaluating intermediate steps \cite{cobbe2021verifiers,uesato2022processoutcome,lightman2023verify}. Multimodal PRMs such as VisualPRM, VRPRM, and MM-PRM extend this idea to visually grounded reasoning \cite{wang2025visualprm,chen2025vrprm,du2025mmprm}. Structured and generative reward models provide additional partial credit or explicit verification, but generally require a learned evaluator or extra model inference during policy optimization \cite{zhang2025structvrm,khalifa2025thinkprm,zhao2025genprm}.

These alternatives create a \textbf{sparse supervision trap}: outcome rewards are cheap but collapse a complete trajectory into one bit, whereas learned process rewards are informative but computationally expensive. An underused resource is that existing process-supervision datasets already contain validated intermediate steps. Such annotations are typically used to train standalone verifiers, rather than directly reused as policy-optimization signals.

We introduce \textbf{StructReward}, a framework that converts existing validated reference steps into reference-conditioned reinforcement signals. For each training problem, we select one reference trajectory and retain the steps labeled correct by VisualPRM. The policy response is parsed into an ordered sequence of reasoning steps. Local step results are compared with the validated reference results using lightweight numerical, symbolic, and lexical rules. Matched steps receive positive labels, unmatched steps receive zero, and the labels are averaged into a response-level process reward. A gated objective combines this signal with format validity, final-answer correctness, and consistency between the last reasoning result and the reported answer.

This design separates four properties that a terminal reward conflates: whether the response is parseable, whether its final answer is correct, whether its intermediate results agree with a validated trajectory, and whether the reported answer is consistent with the last local result. The separation makes each reward component deterministic and auditable. It also keeps the online optimization loop independent of a learned critic or an external language-model judge.

StructReward does not attempt to learn a general-purpose semantic verifier. Its reward determines whether generated local results agree with validated results for the same training problem. This narrower scope avoids an additional verifier forward pass, but requires correct reference steps and does not establish the semantic validity of an arbitrary derivation. A valid alternative derivation may expose different intermediate results and therefore receive incomplete process credit. Accordingly, we use reference agreement as a training signal rather than as proof that a generated chain of thought is uniquely correct or semantically faithful.

We also reuse the responses already generated during policy sampling. After the GRPO update, these rollouts become outcome-judgment and pairwise-selection examples supervised solely by final-answer correctness. In parallel, a strong language model rewrites exactly one step in a current-policy answer-correct trajectory. The original and perturbed trajectories yield error-detection, localization, and correct-version-selection examples. The strong model constructs auxiliary training data only; it is never used to judge rollouts or compute online rewards.

Our contributions are:
\begin{itemize}
    \item We introduce a \textbf{lightweight step-reward mechanism} that transfers validated VisualPRM step labels to policy rollouts through monotonic numerical, symbolic, and lexical result matching.
    \item We \textbf{recycle sampled policy responses} into outcome-judgment and pairwise-selection supervision without additional rollout generation.
    \item We construct \textbf{controlled single-step hard negatives} with a strong language model to improve error detection, localization, and correct-version discrimination.
\end{itemize}

\section{Related Work}

\subsection{Multimodal Reasoning with Verifiable Rewards}

Multimodal large language models have progressed from direct visual question answering toward explicit reasoning over images, diagrams, and documents \cite{wang2025mcotsurvey,zhou2025reinforcedmllm,zhang2026posttraining,bai2025qwen3vl}. Broader benchmark efforts such as UniM also evaluate unified any-to-any interleaved multimodal capabilities \cite{li2026unim}. RLVR has become a central post-training mechanism because many reasoning tasks expose answers that can be checked deterministically \cite{shao2024deepseekmath,deepseekai2025r1}. Its multimodal extensions use structured cold starts, rule-based visual rewards, curricula, staged optimization, and step-wise policy updates \cite{huang2025visionr1,zhan2025visionr1,meng2025mmeureka,wei2025coldstart,chen2025revisualr1,zhang2025r1vl,jeddi2025pcgrpo}. Other methods alternate supervised learning and RL or use trajectory-guided supervision to improve data efficiency \cite{deng2025openvlthinker,wang2025sotawithless,lu2026tgrl}.

Most of these methods still rely primarily on terminal correctness. Outcome rewards scale well, but provide limited resolution for intermediate credit assignment \cite{uesato2022processoutcome,lightman2023verify,setlur2024rewardingprogress,cui2025prime}. StructReward retains the group-relative optimization paradigm while augmenting the outcome reward with a same-problem signal derived from validated intermediate results.

\subsection{Process Reward Models and Multimodal Critics}

Process supervision assigns credit to intermediate steps and can improve error detection, candidate selection, and policy optimization \cite{cobbe2021verifiers,uesato2022processoutcome,lightman2023verify}. Later work reduces annotation cost through automatically generated labels, rollout-based value estimation, and implicit process reinforcement \cite{wang2023mathshepherd,setlur2024rewardingprogress,cui2025prime}. These studies also show that process labels depend on the continuation policy, trajectory distribution, and evaluation protocol \cite{zhang2025prmlessons}.

VisualPRM provides large-scale multimodal process annotations, while MM-PRM scales step-level supervision using generated trajectories \cite{wang2025visualprm,du2025mmprm}. VRPRM and related methods explicitly reason before assigning step scores \cite{chen2025vrprm,khalifa2025thinkprm,she2025rprm}. Fine-grained multimodal supervision further highlights the importance of step decomposition, perceptual grounding, label construction, and evaluation reliability \cite{chen2025chainofstep,ong2025vlprmlessons}.

Generative and trajectory-aware PRMs use verification chains or future rollouts to assess candidates \cite{zhao2025genprm,zou2025reasonflux,wu2026processthinker}. GM-PRM extends process evaluation toward active correction \cite{zhang2025gmprm}. StructVRM provides component-level rewards for structured answers \cite{zhang2025structvrm}. General multimodal critics learn from preferences, critiques, or fine-grained alignment data \cite{xiong2024llavacritic,zang2025ixcreward,zhang2025mmrlhf}. Other work formulates reward modeling itself as a reasoning task \cite{chen2025rmr1,wang2025unifiedreward,zhang2025r1reward,wang2026msrl}, with dedicated benchmarks evaluating perception, hallucination, knowledge, preference alignment, and reasoning \cite{li2024vlrewardbench,yasunaga2025multimodalrewardbench}.

Unlike these general evaluators, StructReward uses a fixed, validated trajectory for the same problem and performs rule-based result matching. It therefore removes learned-critic inference from the GRPO loop, while accepting a narrower reference-conditioned notion of step correctness.

\subsection{Rollout Recycling and Error-Aware Training}

Policy rollouts can provide reusable supervision beyond a single policy-gradient update. Iterative pipelines alternate supervised learning and RL, while policy--reward co-evolution methods reorganize online responses into judgment or reflection objectives \cite{deng2025openvlthinker,liu2025spark}. Perturbed rollout conditions, selected trajectories, and stronger-model guidance can further improve exploration and data efficiency \cite{liu2025noisyrollout,wang2025sotawithless,lu2026tgrl}.

SPARK is closely related because it converts online responses into pointwise judgment, pairwise comparison, and reflection-oriented objectives \cite{liu2025spark}. Preference pairs, textual critiques, and critique-before-scoring objectives also improve judgment quality \cite{xiong2024llavacritic,zhang2025mmrlhf,yu2024criticrm}. StructReward treats rollout recycling as auxiliary supervised training of the policy itself rather than as joint training of a separate reward model.

Critique and reflection provide complementary mechanisms for learning from failures \cite{bai2022constitutional,yu2024criticrm}. Recent systems optimize self-verification, reflection-aware generation, and trajectory-level correction \cite{ma2025s2r,wan2025srpo,wang2025vlrethinker,ding2025sherlock}. Visually grounded reflection emphasizes that correction should remain tied to image evidence \cite{tang2026vgrreflection}. StructReward contributes controlled, localized negatives: exactly one step changes, the remaining trajectory is preserved, and the strong model is excluded from reward computation.

\section{Method}

\subsection{Problem Formulation}

Let a multimodal reasoning problem be \(x=(I,P)\), where \(I\) is an image, \(P\) is the textual prompt, and \(\hat{a}\) is the verified answer. The policy \(\pi_\theta\) generates
\begin{equation}
o=(s_1,s_2,\ldots,s_M,a_{\mathrm{out}}),
\end{equation}
where \(s_i=(c_i,a_i)\) contains reasoning text \(c_i\) and a parsed local result \(a_i\). StructReward compares local results rather than attempting to judge the complete semantics of \(c_i\).

Local results provide a tractable interface between free-form reasoning and deterministic reward computation. Numerical answers can be compared after normalization, symbolic expressions can be canonicalized, and short textual answers can be matched after standardization. The surrounding explanation remains part of the policy response, but it is not assigned a semantic score by the rule-based matcher. This choice deliberately limits the reward to properties that can be reproduced without a learned evaluator.

\begin{figure*}[!t]
\centering
\includegraphics[width=0.96\textwidth]{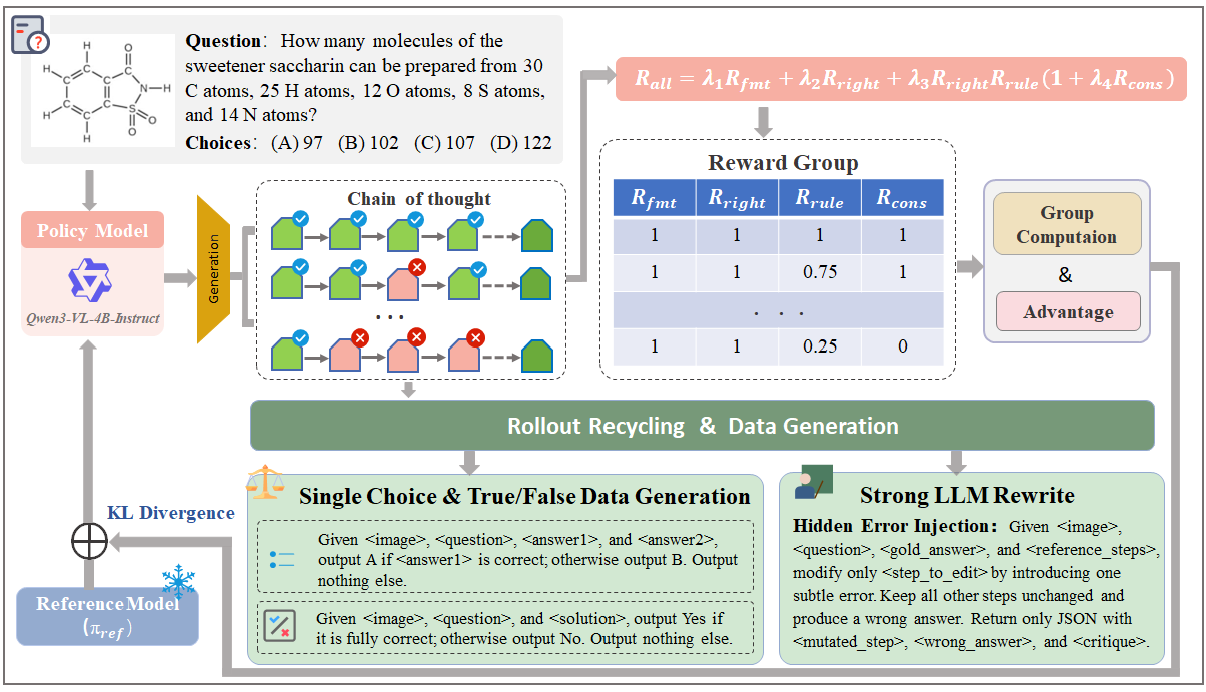}
\caption{Overview of StructReward. Structured process rewards guide policy optimization, while policy rollouts are recycled into judgment and selection data and a strong language model constructs localized hard negatives.}
\label{fig:framework}
\end{figure*}

For every training problem, preprocessing selects one VisualPRM reference trajectory and retains only its correct steps:
\begin{equation}
\tau^+=\{(s_j^+,a_j^+)\}_{j=1}^{N}.
\end{equation}
This reference remains fixed for all rollouts of the problem. The training pipeline comprises cold-start supervised fine-tuning (SFT), structured-reward GRPO, rollout recycling, and online hard-negative construction. Figure~\ref{fig:framework} summarizes the complete procedure. Reference trajectories and matching rules are used only during training; inference requires neither references nor a reward model.

\subsection{Cold-Start Supervised Fine-Tuning}

\paragraph{Data selection.}
We select 5,000 long-form examples from VisualPRM400K-v1.1 \cite{wang2025visualprm}. Each example contains an image--question pair, a step-by-step solution, and step labels. Samples contain 3--12 reasoning steps and span science, geometry, functions, physics, and biology. Selection balances source subsets and ranks candidates using reasoning length (0.42), number of steps (0.23), visual dependence (0.20), label quality (0.10), and format validity (0.05).

\paragraph{Format normalization.}
Following evidence that a structured cold start stabilizes multimodal RL \cite{huang2025visionr1,wei2025coldstart,chen2025revisualr1}, every target contains exactly one \texttt{<think>}\ldots\texttt{</think>} block. Steps begin with consecutive \texttt{Step i:} markers and contain an extractable local result expressed with \texttt{\textbackslash boxed\{\}}. A separate boxed expression outside the reasoning block reports the final answer.

The normalized format is not treated as evidence of reasoning quality. Instead, it defines a stable interface between generation and reward computation: step boundaries expose the units to be aligned, local boxes expose the values to be compared, and the separate answer field permits an independent correctness check. Malformed outputs can therefore be identified before any step matching is attempted.

\paragraph{Objective.}
For target response \(o^*\), cold-start training minimizes
\begin{equation}
\mathcal{L}_{\mathrm{SFT}}
=-\mathbb{E}_{(x,o^*)\sim\mathcal{D}_{\mathrm{SFT}}}
\left[\sum_{t=1}^{|o^*|}
\log \pi_\theta(o_t^*\mid x,o_{<t}^*)\right].
\end{equation}

\subsection{Structured Reward Design}

\subsubsection{Format Reward}
The parser requires exactly one reasoning block, consecutive step indices starting at 1, a local result in every step, and a separate final answer. The format reward is
\begin{equation}
R_{\mathrm{fmt}}=
\begin{cases}
1,&\text{if all structural constraints are satisfied},\\
0,&\text{otherwise}.
\end{cases}
\end{equation}

\subsubsection{Step-Result Matching}
For each generated result \(a_i\), matching uses the appropriate rule for its type. Numerical values are normalized and compared within a tolerance. Symbolic expressions are canonicalized before equivalence checking. Textual results are normalized for capitalization, whitespace, and standard answer forms.

Let
\begin{equation}
\match(a_i,a_j^+)\in\{0,1\}.
\end{equation}
Generated steps are processed in order and aligned monotonically to reference steps; a matched reference step cannot be reused. If \(\mathcal{A}_i\) is the set of still-available reference indices at step \(i\), then
\begin{equation}
z_i=\ind\!\left[\exists j\in\mathcal{A}_i:
\match(a_i,a_j^+)=1\right].
\end{equation}
The response-level process reward is
\begin{equation}
R_{\mathrm{rule}}=\frac{1}{M}\sum_{i=1}^{M} z_i.
\end{equation}
Unmatched or redundant steps contribute zero, so adding irrelevant steps cannot increase the reward. The quantity measures agreement with reference results, not general semantic correctness.

Monotonic one-to-one alignment preserves the order of the validated computation and prevents reward inflation through repetition. It is intentionally more conservative than unconstrained set matching: once a reference result has been consumed, later copies cannot claim the same credit. At the same time, matching local results rather than exact reasoning text permits lexical variation when the normalized numerical or symbolic result is equivalent.

\subsubsection{Consistency and Correctness}
Let \(a_{\mathrm{reasoned}}\) be the last parseable local result. We define
\begin{align}
R_{\mathrm{cons}}&=\ind[a_{\mathrm{reasoned}}\equiv a_{\mathrm{out}}],\\
R_{\mathrm{right}}&=\ind[a_{\mathrm{out}}\equiv\hat{a}],
\end{align}
where \(\equiv\) denotes equivalence after unified parsing and normalization. The consistency term constrains the generated output; it is not evidence that the internal reasoning is faithful.

\subsubsection{Gated Total Reward}
The total reward is
\begin{equation}
\begin{split}
R_{\mathrm{total}}={}&
\lambda_{\mathrm{fmt}}R_{\mathrm{fmt}}
+\lambda_{\mathrm{right}}R_{\mathrm{right}}\\
&+\lambda_{\mathrm{rule}}R_{\mathrm{right}}R_{\mathrm{rule}}
(1+\lambda_{\mathrm{cons}}R_{\mathrm{cons}}),
\end{split}
\end{equation}
where \(\lambda_{\mathrm{fmt}}=0.10\), \(\lambda_{\mathrm{right}}=0.25\),
\(\lambda_{\mathrm{rule}}=0.75\), and \(\lambda_{\mathrm{cons}}=0.15\).
Final-answer correctness gates the process term. Consequently, partial reference matches cannot compensate for a wrong final answer, while the process reward differentiates among answer-correct trajectories.

Each term has a distinct role. \(R_{\mathrm{fmt}}\) rewards compliance with the parsing contract; \(R_{\mathrm{right}}\) preserves the verifiable task objective; \(R_{\mathrm{rule}}\) supplies resolution among trajectories that reach the correct answer; and \(R_{\mathrm{cons}}\) checks agreement between the final local result and the answer field. The multiplicative gate is essential to the intended interpretation: step overlap is auxiliary evidence only after terminal correctness has been established.

\subsection{Stable Group-Relative Policy Optimization}

For each problem, the policy samples \(G=8\) responses. The normalized advantage for response \(i\) is
\begin{equation}
A_i=\clip\!\left(
\frac{R_i-\bar{R}}{\sigma_R+10^{-8}},-3,3\right).
\end{equation}
When \(\sigma_R<10^{-4}\), the group has no meaningful relative signal and is excluded from the update. This prevents flat reward groups from producing unstable normalization.

The exclusion criterion also avoids assigning artificial preferences when every sampled response receives effectively the same reward. For non-flat groups, standardization makes the update depend on within-problem reward differences rather than on the absolute scale of rewards across heterogeneous problems, and clipping limits the influence of unusually large normalized advantages.

Let \(\pi_{\mathrm{old}}\) be the rollout policy. The token-level ratio is
\begin{equation}
\begin{split}
\rho_{i,t}(\theta)=\exp\bigl[
&\log\pi_\theta(o_{i,t}\mid x,o_{i,<t})\\
&-\log\pi_{\mathrm{old}}(o_{i,t}\mid x,o_{i,<t})
\bigr].
\end{split}
\end{equation}
We minimize the clipped GRPO objective \cite{shao2024deepseekmath}:
\begin{equation}
\begin{split}
\mathcal{L}_{\mathrm{GRPO}}={}&-\frac{1}{G}
\sum_{i=1}^{G}\frac{1}{|o_i|}
\sum_{t=1}^{|o_i|}
\min\Big(
\rho_{i,t}A_i,\\
&\clip(\rho_{i,t},1-\epsilon,1+\epsilon)A_i
\Big)
+\beta D_{\mathrm{KL}}(\pi_\theta\|\pi_{\mathrm{ref}}),
\end{split}
\end{equation}
where \(\pi_{\mathrm{ref}}\) is the frozen cold-start policy.

\subsection{Rollout Recycling}

After the GRPO update, the same responses become auxiliary supervised examples. The outcome label is
\begin{equation}
y_i=\ind[a_{\mathrm{out}}^{(i)}\equiv\hat{a}].
\end{equation}
In the pointwise task, the model generates \texttt{Correct} or \texttt{Incorrect} for one candidate response. In the pairwise task, one answer-correct and one answer-incorrect response from the same rollout group are randomly assigned to A and B, and the model generates the label of the correct candidate. Candidate order is randomized independently, and pairs with identical outcome labels are excluded.

The two tasks expose complementary views of the same sampled data. Pointwise judgment asks whether one trajectory reaches the verified answer, whereas pairwise selection asks the model to discriminate between two trajectories for the same problem. Because both labels are derived from deterministic final-answer checking, recycling introduces no additional judge labels and requires no extra policy rollouts.

Let
\(\mathcal{D}_{\mathrm{recycle}}=\{(u_n,v_n)\}_{n=1}^{N_r}\).
The auxiliary target-sequence loss is
\begin{equation}
\mathcal{L}_{\mathrm{recycle}}=
-\mathbb{E}_{(u,v)\sim\mathcal{D}_{\mathrm{recycle}}}
\left[\sum_{t=1}^{|v|}
\log\pi_\theta(v_t\mid u,v_{<t})\right].
\end{equation}

\subsection{Online Hard-Negative Construction}

Naturally incorrect rollouts may contain several interacting errors. We therefore select format-valid, answer-correct current-policy trajectories and ask a strong language model to rewrite exactly one step. The content, numbering, and order of all other steps remain unchanged. Deterministic checks retain a sample only if exactly one step changed materially and the complete response remains parseable.

Let \(o^+\) be the original, \(o^-\) the perturbed trajectory, and \(j^*\) the changed step. Each pair produces an error-detection and localization task, plus a correct-version-selection task with randomized candidate order. For
\(\mathcal{D}_{\mathrm{hard}}=\{(\tilde{u}_n,\tilde{v}_n)\}_{n=1}^{N_h}\),
the loss is
\begin{equation}
\mathcal{L}_{\mathrm{hard}}=
-\mathbb{E}_{(\tilde{u},\tilde{v})\sim\mathcal{D}_{\mathrm{hard}}}
\left[\sum_{t=1}^{|\tilde{v}|}
\log\pi_\theta(\tilde{v}_t\mid\tilde{u},\tilde{v}_{<t})\right].
\end{equation}
The strong model is used exclusively for constructing auxiliary data. It neither scores policy responses nor participates in online reward computation.

Localizing the perturbation to one step reduces ambiguity about the supervision target: the original and rewritten responses share the same problem, context, step count, and unchanged surrounding steps. Deterministic filtering enforces these structural conditions before a pair is admitted. This auxiliary construction is distinct from the reference-conditioned reward: the former creates supervised error-awareness examples, whereas the latter directly supplies reward during GRPO.

\begin{table*}[!t]
\centering
\setlength{\tabcolsep}{4.2pt}
\renewcommand{\arraystretch}{1.10}
\arrayrulecolor{RuleGray}
\small
\begin{tabular}{
    >{\raggedright\arraybackslash}p{3.2cm}
    !{\color{RuleGray}\vrule width 0.65pt}
    cc
    cc
    !{\color{RuleGray}\vrule width 0.65pt}
    c
}
\toprule[1.15pt]
\rowcolor{HeaderGray}
\multirow{2}{*}{\textbf{Method}} &
\multicolumn{2}{c}{\textbf{General Multimodal Reasoning}} &
\multicolumn{2}{c}{\textbf{Visual Mathematical Reasoning}} &
\multirow{2}{*}{\textbf{Reported Avg.}} \\
\cmidrule(lr){2-3}\cmidrule(lr){4-5}
\rowcolor{HeaderGray}
& \textbf{MMMU} & \textbf{MMMU-Pro}
& \textbf{MathVista$_{\mathrm{mini}}$}
& \textbf{MathVerse$_{\mathrm{mini}}$} & \\
\midrule[0.85pt]
Qwen3-VL-2B-Instruct & 53.4 & 36.5 & 61.3 & 52.1 & 50.8 \\
\rowcolor{HighlightGray}
\textbf{StructReward-2B} & \textbf{57.7} & \textbf{40.2} & \textbf{64.1} & \textbf{55.7} & \textbf{54.4} \\
\rowcolor{LightGray}
\textit{Gain over Instruct} & \textit{+4.3} & \textit{+3.7} & \textit{+2.8} & \textit{+3.6} & \textit{+3.6} \\
\midrule[0.55pt]
Qwen3-VL-4B-Instruct & 67.4 & 53.2 & 73.7 & 46.8 & 60.3 \\
\rowcolor{HighlightGray}
\textbf{StructReward-4B} & \textbf{70.3} & \textbf{56.7} & \textbf{81.0} & \textbf{49.6} & \textbf{64.4} \\
\rowcolor{LightGray}
\textit{Gain over Instruct} & \textit{+2.9} & \textit{+3.5} & \textit{+7.3} & \textit{+2.8} & \textit{+4.1} \\
\midrule[0.55pt]
Qwen3-VL-8B-Instruct & 69.6 & 55.9 & 77.2 & 62.1 & 66.2 \\
\rowcolor{HighlightGray}
\textbf{StructReward-8B} & \textbf{74.2} & \textbf{60.3} & \textbf{82.2} & \textbf{65.8} & \textbf{70.6} \\
\rowcolor{LightGray}
\textit{Gain over Instruct} & \textit{+4.6} & \textit{+4.4} & \textit{+5.0} & \textit{+3.7} & \textit{+4.4} \\
\bottomrule[1.15pt]
\end{tabular}
\caption{Main results across Qwen3-VL backbone scales.}
\label{tab:main-results}
\end{table*}

\section{Experiments}

\subsection{Experimental Setup}

\paragraph{Benchmarks.}
We evaluate general multimodal reasoning on MMMU and MMMU-Pro \cite{yue2023mmmu,yue2024mmmupro}. MMMU covers college-level knowledge across many disciplines; MMMU-Pro increases perceptual and reasoning difficulty with harder questions and distractors. Visual mathematical reasoning is evaluated on MathVista, MATH-Vision, and MathVerse \cite{lu2023mathvista,wang2024mathvision,zhang2024mathverse}. MathVista covers diagrams, charts, and visual contexts; MATH-Vision contains competition-level visual mathematics; and MathVerse emphasizes genuine dependence on visual information. We follow the official evaluation protocol for each benchmark and report answer accuracy.

\paragraph{Training data.}
All scales use the same 5,000-example SFT corpus selected from VisualPRM400K-v1.1 \cite{wang2025visualprm}. We retain trajectories with valid images, verified answers, and well-formed steps, while removing malformed samples, inaccessible images, near duplicates, answer-only responses, and trajectories whose final answers disagree with verified labels. The structured-RL prompt pool is disjoint from SFT and evaluation data. Overlap is removed using normalized question identifiers, question hashes, image paths, and image hashes.

\paragraph{Models.}
We use Qwen3-VL-2B-Instruct, Qwen3-VL-4B-Instruct, and Qwen3-VL-8B-Instruct as backbones \cite{bai2025qwen3vl}. StructReward applies the shared SFT corpus, structured RL, rollout recycling, and hard-negative training. The cross-scale comparison tests whether gains transfer across capacities; it is not a controlled scaling-law study because training compute varies with model size.

\paragraph{Optimization.}
For each problem, we sample \(G=8\) rollouts. The ratio-clipping parameter is 0.2, normalized advantages are clipped to \([-3,3]\), and the reference-policy KL coefficient is \(10^{-3}\). All three model sizes---2B, 4B, and 8B---use 1,000 prompt groups, corresponding to 8,000 rollouts per run. Auxiliary pointwise and pairwise tasks use autoregressive target-sequence cross-entropy rather than a classification head.

\paragraph{Reporting protocol.}
We report the point estimates recorded for the evaluated checkpoints and do not report confidence intervals or multi-seed variation. Consequently, the benchmark-level accuracies and their direct differences are the primary empirical evidence; we do not interpret small gaps as statistically significant. The displayed cross-scale results are not compute-matched because training compute varies with model size even though the prompt-group budgets are identical. These reporting boundaries keep the analysis aligned with the available experiments.

\paragraph{Average calculation.}
For every row in the main-results and ablation tables, ``Reported Avg.'' is the arithmetic mean of the four displayed benchmarks---MMMU, MMMU-Pro, MathVista, and MathVerse---rounded to one decimal place. Gain values are computed from these displayed four-benchmark averages.

\subsection{Main Results}

Table~\ref{tab:main-results} reports accuracy for all three model scales. StructReward improves every displayed benchmark result. On Qwen3-VL-2B, the four-benchmark average rises from 50.8 to 54.4, an absolute gain of 3.6 points. The gains range from 2.8 points on MathVista to 4.3 points on MMMU.

On Qwen3-VL-4B, StructReward improves MMMU and MMMU-Pro by 2.9 and 3.5 points. Its largest displayed gain is on MathVista, where accuracy rises from 73.7 to 81.0. MathVerse improves by 2.8 points, and the four-benchmark average increases by 4.1 points.

On Qwen3-VL-8B, StructReward improves MMMU by 4.6 points, MMMU-Pro by 4.4, MathVista by 5.0, and MathVerse by 3.7. The 4.4-point increase in the four-benchmark average indicates that the framework remains effective for a stronger backbone.

Across the entries shown in Table~\ref{tab:main-results}, no benchmark reverses the direction of the comparison: every StructReward value exceeds its corresponding instruction-tuned baseline. The magnitude is not uniform across tasks or scales, which is expected because the benchmarks emphasize different mixtures of perception, domain knowledge, and mathematical reasoning. The table establishes an empirical association between the complete training procedure and higher checkpoint accuracy; it does not, by itself, isolate which reward component causes each benchmark-specific gain.

\begin{table*}[!t]
\centering
\setlength{\tabcolsep}{3.0pt}
\renewcommand{\arraystretch}{1.14}
\arrayrulecolor{RuleGray}
\small
\begin{tabular}{
    >{\raggedright\arraybackslash}p{3.7cm}
    !{\color{RuleGray}\vrule width 0.65pt}
    ccc
    !{\color{RuleGray}\vrule width 0.65pt}
    cc
    cc
    !{\color{RuleGray}\vrule width 0.65pt}
    c
}
\toprule[1.15pt]
\rowcolor{HeaderGray}
\multirow{2}{*}{\textbf{Variant}} &
\multicolumn{3}{c}{\textbf{Components}} &
\multicolumn{2}{c}{\textbf{General Multimodal Reasoning}} &
\multicolumn{2}{c}{\textbf{Visual Mathematical Reasoning}} &
\multirow{2}{*}{\textbf{Reported Avg.}} \\
\cmidrule(lr){2-4}\cmidrule(lr){5-6}\cmidrule(lr){7-8}
\rowcolor{HeaderGray}
& \textbf{SR} & \textbf{RW} & \textbf{JC}
& \textbf{MMMU} & \textbf{MMMU-Pro}
& \textbf{MathVista} & \textbf{MathVerse} & \\
\midrule[0.85pt]
Qwen3-VL-4B-Instruct & \xmark & \xmark & \xmark & 67.4 & 53.2 & 73.7 & 46.8 & 60.3 \\
\rowcolor{LightGray}
w/o strong-LLM rewriting & \cmark & \xmark & \cmark & \underline{69.4} & \underline{54.9} & 78.7 & 48.8 & 63.0 \\
w/o judgment/choice generation & \cmark & \cmark & \xmark & 68.9 & 54.7 & \underline{80.8} & \underline{49.5} & \underline{63.5} \\
\rowcolor{HighlightGray}
\textbf{Full StructReward} & \cmark & \cmark & \cmark & \textbf{70.3} & \textbf{56.7} & \textbf{81.0} & \textbf{49.6} & \textbf{64.4} \\
\bottomrule[1.15pt]
\end{tabular}
\caption{Ablation study on Qwen3-VL-4B. SR, RW, and JC denote structured RL, strong-LLM rewriting, and judgment/choice generation, respectively.}
\label{tab:ablation}
\end{table*}

\begin{figure}[!t]
\centering
\includegraphics[width=0.74\columnwidth]{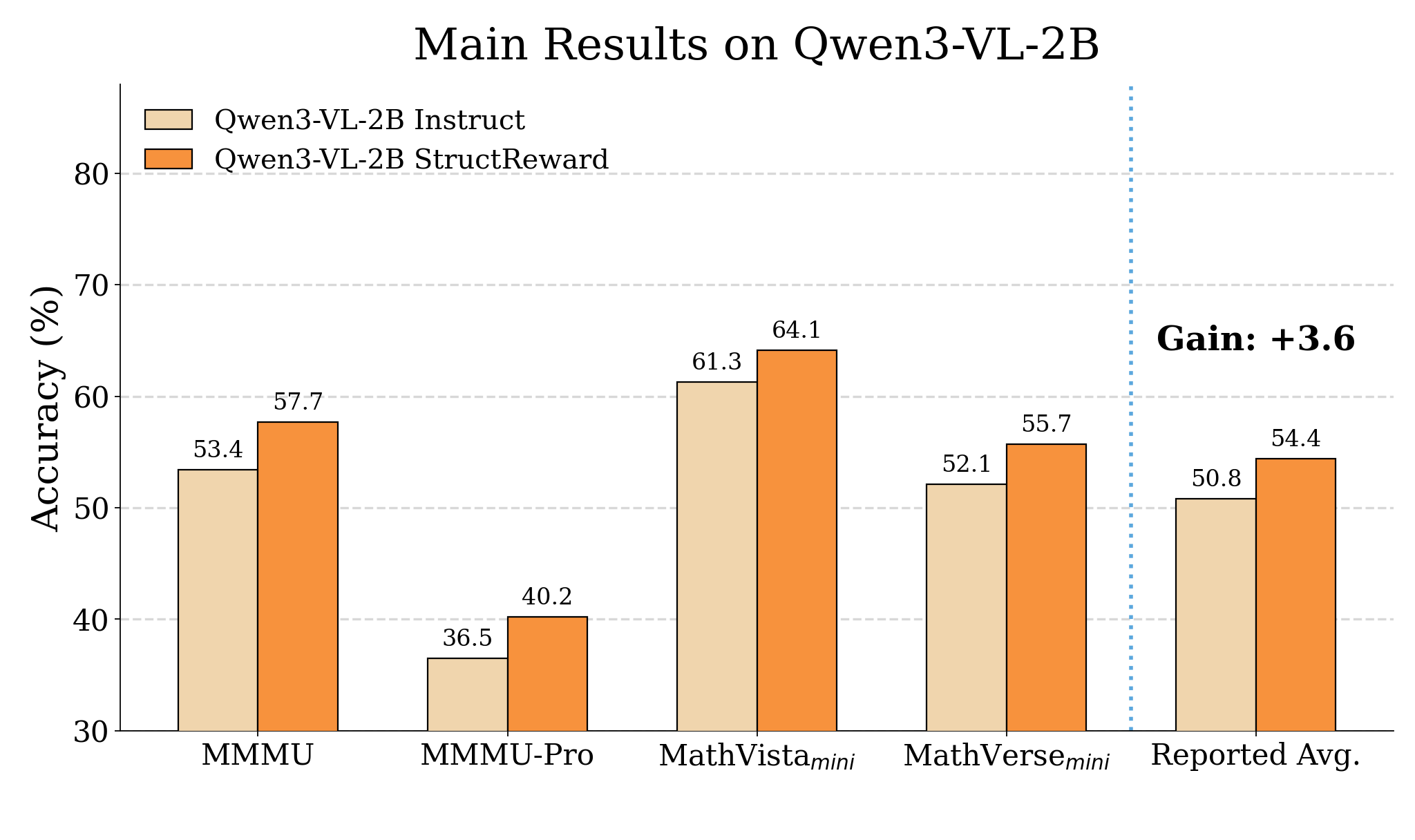}\\[-0.35em]
\footnotesize (a) Qwen3-VL-2B: gains across all benchmarks.\\[0.25em]
\includegraphics[width=0.74\columnwidth]{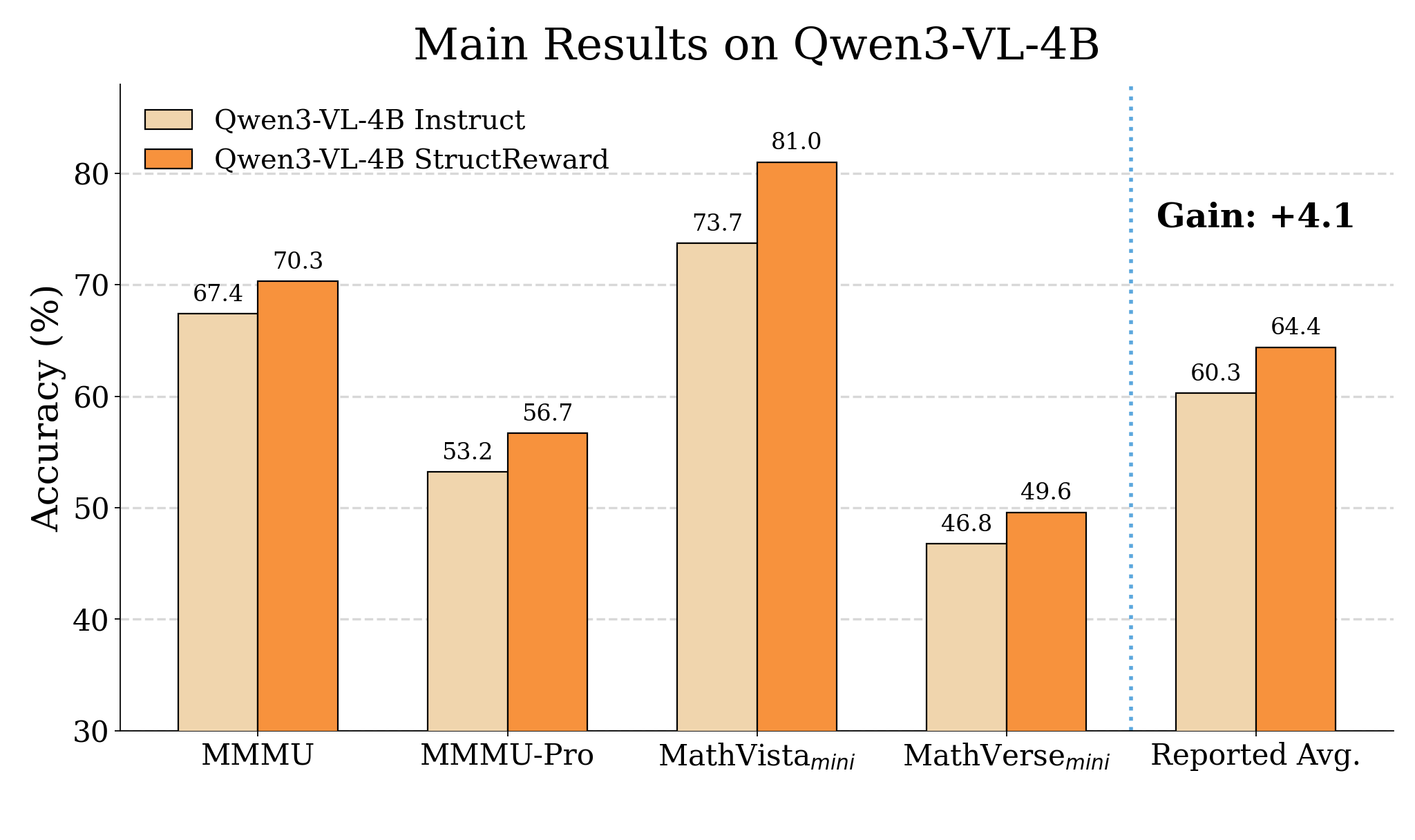}\\[-0.35em]
\footnotesize (b) Qwen3-VL-4B: strongest gain on MathVista.\\[0.25em]
\includegraphics[width=0.74\columnwidth]{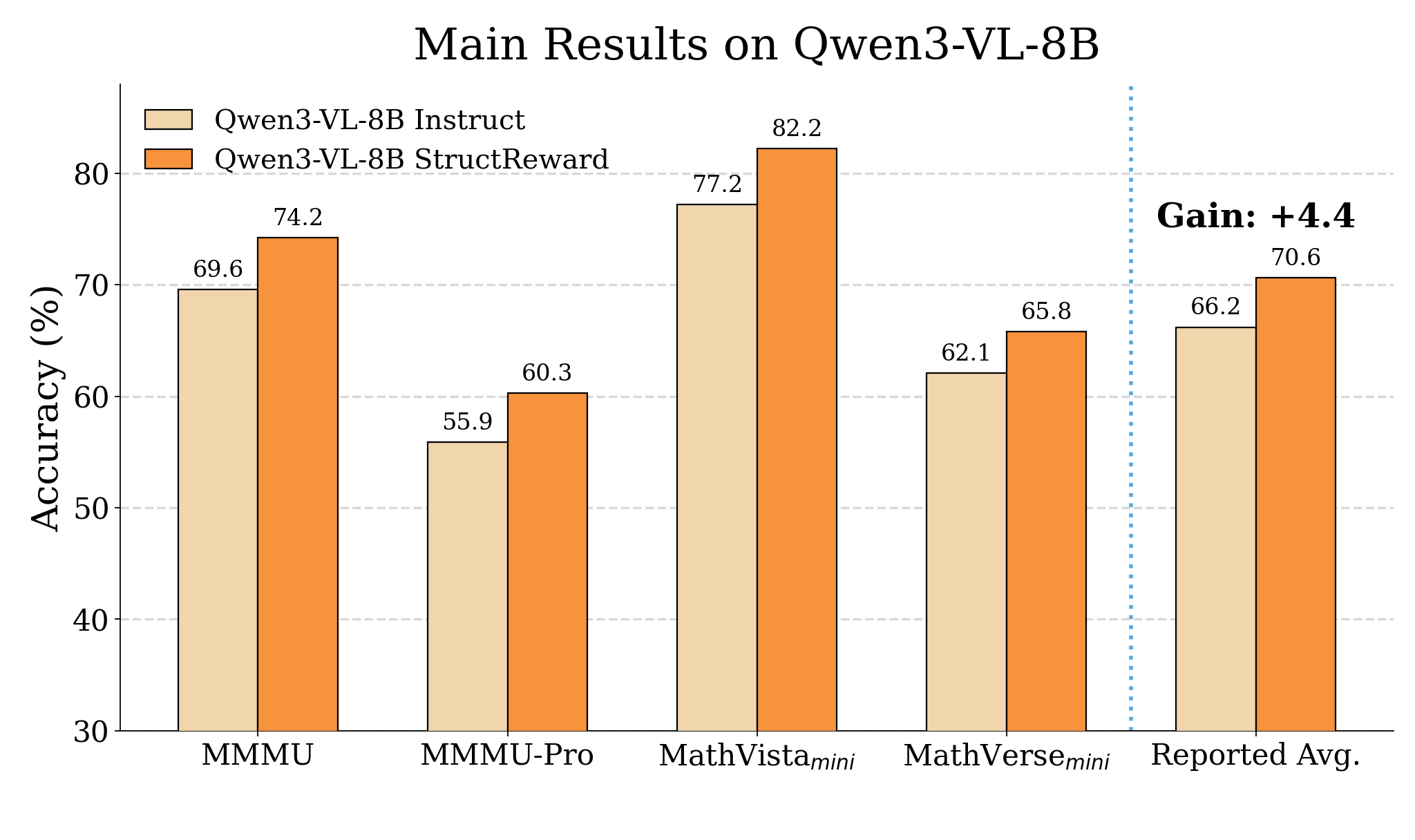}\\[-0.35em]
\footnotesize (c) Qwen3-VL-8B: improvements persist at scale.
\caption{Single-column comparison of Qwen3-VL-Instruct and StructReward across the three backbone scales.}
\label{fig:main-results}
\end{figure}

\paragraph{Effect of scale.}
The gains are not confined to one capacity. Improvements on MMMU-Pro and MathVista remain substantial as model size increases, suggesting that reference-conditioned process supervision complements reasoning ability already present in the backbone. Because all scales use identical prompt-group budgets but different backbone sizes, Figure~\ref{fig:main-results} provides a budget-aligned comparison across capacities rather than a compute-matched scaling law.

\FloatBarrier

\subsection{Ablation Study}

All ablations use Qwen3-VL-4B and the same SFT initialization, structured-RL prompts, rollout budget, optimization settings, decoding configuration, and evaluation examples. The reduced variant \emph{w/o strong-LLM rewriting} retains structured RL and rollout-derived judgment/choice supervision but removes single-step hard negatives. The variant \emph{w/o judgment/choice generation} retains structured RL and strong-model rewriting but removes outcome-judgment and pairwise-choice examples. Full StructReward includes all three components.

Table~\ref{tab:ablation} shows that both reduced variants improve substantially over Qwen3-VL-4B-Instruct. Removing strong-model rewriting lowers the four-benchmark average from 64.4 to 63.0, while removing rollout-derived judgment and choice lowers it to 63.5. The full framework achieves the best result on every displayed benchmark. These comparisons indicate that localized hard negatives and rollout recycling provide complementary supervision.

Relative to the full configuration, removing rewriting produces a 1.4-point decrease in the four-benchmark average, and removing judgment/choice generation produces a 0.9-point decrease. These differences are computed directly from the displayed values. Because the study removes one auxiliary component at a time and reports single checkpoint estimates, the comparison should be read as component evidence within this training configuration rather than as a complete factorial or statistical decomposition of interactions.

\begin{figure}[!t]
\centering
\includegraphics[width=\columnwidth]{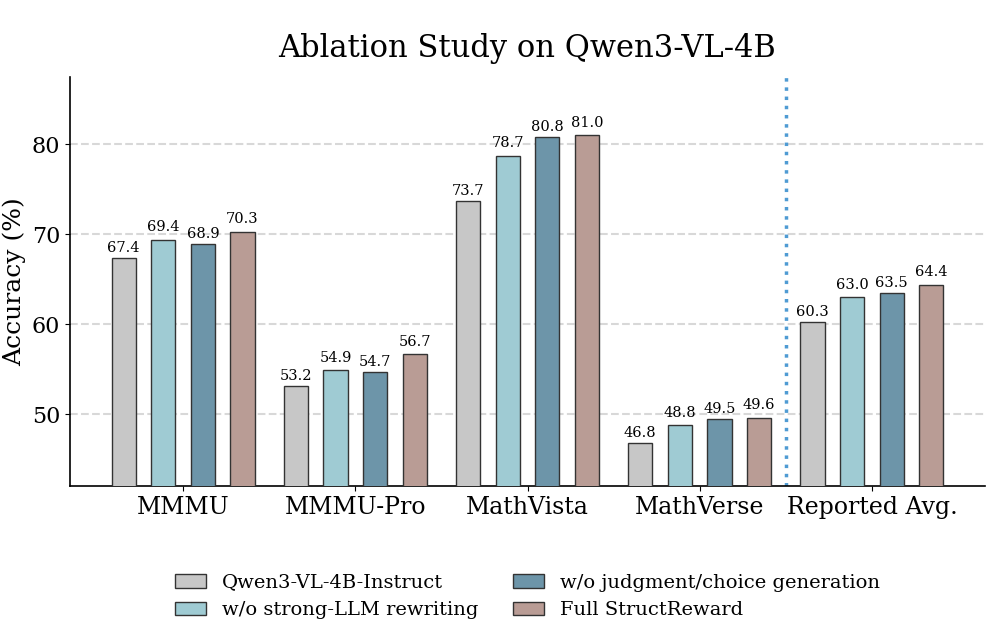}
\caption{Qwen3-VL-4B ablations; rewriting and rollout recycling provide complementary gains.}
\label{fig:ablation}
\end{figure}

\section{Discussion}

\paragraph{Interpretation of the reward.}
StructReward supervises agreement with a validated trajectory for the same problem, not unrestricted semantic verification. This distinction matters when several derivations are valid: a rollout can reach the verified answer through intermediate quantities that are absent from the selected reference. Final-answer gating prevents a partial reference match from overriding an incorrect outcome, but it cannot make the selected reference exhaustive. The reward is therefore best understood as conservative, reference-conditioned partial credit.

\paragraph{Evidence and computational scope.}
The online reward path uses parsing, normalization, equivalence checks, and aggregation; the strong language model is confined to auxiliary data construction. This supports the claim of lightweight online reward computation, but we do not report hardware-matched timing and therefore do not claim a measured end-to-end speedup. The results cover three Qwen3-VL capacities, while the ablation is limited to the 4B setting. Multiple seeds, additional model families, and comparisons among valid references would provide evidence beyond the point estimates reported here.

\section{Conclusion}

We presented StructReward, which converts validated intermediate results into structured reinforcement signals without a learned online verifier. Its gated GRPO reward combines rule-based alignment, answer correctness, reasoning--answer consistency, and format validity. Rollout recycling and controlled single-step perturbations provide complementary auxiliary supervision. Across the reported Qwen3-VL scales, the results show that existing process annotations can be reused as dense, reference-conditioned policy supervision without claiming general semantic verification.

\clearpage
\bibliography{references}

@misc{wang2025mcotsurvey,
  author = {Wang, Yaoting and others},
  title = {Multimodal Chain-of-Thought Reasoning: A Comprehensive Survey},
  year = {2025},
  eprint = {2503.12605},
  archivePrefix = {arXiv}
}

@misc{zhou2025reinforcedmllm,
  author = {Zhou, Guanghao and others},
  title = {Reinforced {MLLM}: A Survey on {RL}-Based Reasoning in Multimodal Large Language Models},
  year = {2025},
  eprint = {2504.21277},
  archivePrefix = {arXiv}
}

@misc{zhang2026posttraining,
  author = {Zhang, Haonan and others},
  title = {A Survey on Post-Training of Multimodal Large Language Models},
  year = {2026},
  note = {Preprints 202607.1494}
}

@misc{shao2024deepseekmath,
  author = {Shao, Zhihong and others},
  title = {{DeepSeekMath}: Pushing the Limits of Mathematical Reasoning in Open Language Models},
  year = {2024},
  eprint = {2402.03300},
  archivePrefix = {arXiv}
}

@misc{bai2025qwen3vl,
  author = {Bai, Shuai and others},
  title = {{Qwen3-VL} Technical Report},
  year = {2025},
  eprint = {2511.21631},
  archivePrefix = {arXiv}
}

@misc{chen2025revisualr1,
  author = {Chen, Shuang and others},
  title = {Advancing Multimodal Reasoning: From Optimized Cold Start to Staged Reinforcement Learning},
  year = {2025},
  eprint = {2506.04207},
  archivePrefix = {arXiv}
}

@misc{deepseekai2025r1,
  author = {{DeepSeek-AI}},
  title = {{DeepSeek-R1}: Incentivizing Reasoning Capability in {LLMs} via Reinforcement Learning},
  year = {2025},
  eprint = {2501.12948},
  archivePrefix = {arXiv}
}

@misc{huang2025visionr1,
  author = {Huang, Wenxuan and others},
  title = {{Vision-R1}: Incentivizing Reasoning Capability in Multimodal Large Language Models},
  year = {2025},
  eprint = {2503.06749},
  archivePrefix = {arXiv}
}

@misc{jeddi2025pcgrpo,
  author = {Jeddi, Ahmadreza and others},
  title = {Puzzle Curriculum {GRPO} for Vision-Centric Reasoning},
  year = {2025},
  eprint = {2512.14944},
  archivePrefix = {arXiv}
}

@misc{meng2025mmeureka,
  author = {Meng, Fanqing and others},
  title = {{MM-Eureka}: Exploring Visual Aha Moment with Rule-Based Large-Scale Reinforcement Learning},
  year = {2025},
  eprint = {2503.07365},
  archivePrefix = {arXiv}
}

@misc{wei2025coldstart,
  author = {Wei, Lai and others},
  title = {Advancing Multimodal Reasoning via Reinforcement Learning with Cold Start},
  year = {2025},
  eprint = {2505.22334},
  archivePrefix = {arXiv}
}

@misc{zhan2025visionr1,
  author = {Zhan, Yufei and others},
  title = {{Vision-R1}: Evolving Human-Free Alignment in Large Vision-Language Models via Vision-Guided Reinforcement Learning},
  year = {2025},
  eprint = {2503.18013},
  archivePrefix = {arXiv}
}

@misc{zhang2025r1vl,
  author = {Zhang, Jingyi and others},
  title = {{R1-VL}: Learning to Reason with Multimodal Large Language Models via Step-Wise Group Relative Policy Optimization},
  year = {2025},
  eprint = {2503.12937},
  archivePrefix = {arXiv}
}

@misc{cobbe2021verifiers,
  author = {Cobbe, Karl and others},
  title = {Training Verifiers to Solve Math Word Problems},
  year = {2021},
  eprint = {2110.14168},
  archivePrefix = {arXiv}
}

@misc{uesato2022processoutcome,
  author = {Uesato, Jonathan and others},
  title = {Solving Math Word Problems with Process- and Outcome-Based Feedback},
  year = {2022},
  eprint = {2211.14275},
  archivePrefix = {arXiv}
}

@misc{lightman2023verify,
  author = {Lightman, Hunter and others},
  title = {Let's Verify Step by Step},
  year = {2023},
  eprint = {2305.20050},
  archivePrefix = {arXiv}
}

@misc{wang2023mathshepherd,
  author = {Wang, Peiyi and others},
  title = {{Math-Shepherd}: Verify and Reinforce {LLMs} Step-by-Step without Human Annotations},
  year = {2023},
  eprint = {2312.08935},
  archivePrefix = {arXiv}
}

@misc{setlur2024rewardingprogress,
  author = {Setlur, Amrith and others},
  title = {Rewarding Progress: Scaling Automated Process Verifiers for {LLM} Reasoning},
  year = {2024},
  eprint = {2410.08146},
  archivePrefix = {arXiv}
}

@misc{chen2025chainofstep,
  author = {Chen, Honghao and Lou, Xingzhou and Feng, Xiaokun and Huang, Kaiqi and Wang, Xinlong},
  title = {Unveiling Chain of Step Reasoning for Vision-Language Models with Fine-Grained Rewards},
  year = {2025},
  eprint = {2509.19003},
  archivePrefix = {arXiv}
}

@misc{chen2025vrprm,
  author = {Chen, Xinquan and Liu, Bangwei and Wang, Xuhong},
  title = {{VRPRM}: Process Reward Modeling via Visual Reasoning},
  year = {2025},
  eprint = {2508.03556},
  archivePrefix = {arXiv}
}

@misc{cui2025prime,
  author = {Cui, Ganqu and others},
  title = {Process Reinforcement through Implicit Rewards},
  year = {2025},
  eprint = {2502.01456},
  archivePrefix = {arXiv}
}

@misc{du2025mmprm,
  author = {Du, Lingxiao and others},
  title = {{MM-PRM}: Enhancing Multimodal Mathematical Reasoning with Scalable Step-Level Supervision},
  year = {2025},
  eprint = {2505.13427},
  archivePrefix = {arXiv}
}

@misc{khalifa2025thinkprm,
  author = {Khalifa, Muhammad and others},
  title = {Process Reward Models That Think},
  year = {2025},
  eprint = {2504.16828},
  archivePrefix = {arXiv}
}

@misc{ong2025vlprmlessons,
  author = {Ong, Brandon and Pala, Tej Deep and Toh, Vernon and Tjhi, William Chandra and Poria, Soujanya},
  title = {Training Vision-Language Process Reward Models for Test-Time Scaling in Multimodal Reasoning: Key Insights and Lessons Learned},
  year = {2025},
  eprint = {2509.23250},
  archivePrefix = {arXiv}
}

@misc{she2025rprm,
  author = {She, Shuaijie and Liu, Junxiao and Liu, Yifeng and Chen, Jiajun and Huang, Xin and Huang, Shujian},
  title = {{R-PRM}: Reasoning-Driven Process Reward Modeling},
  year = {2025},
  eprint = {2503.21295},
  archivePrefix = {arXiv}
}

@misc{wang2025visualprm,
  author = {Wang, Weiyun and others},
  title = {{VisualPRM}: An Effective Process Reward Model for Multimodal Reasoning},
  year = {2025},
  eprint = {2503.10291},
  archivePrefix = {arXiv}
}

@misc{zhang2025gmprm,
  author = {Zhang, Jianghangfan and Yan, Yibo and Zheng, Kening and Zou, Xin and Dai, Song and Hu, Xuming},
  title = {{GM-PRM}: A Generative Multimodal Process Reward Model for Multimodal Mathematical Reasoning},
  year = {2025},
  eprint = {2508.04088},
  archivePrefix = {arXiv}
}

@misc{zhang2025prmlessons,
  author = {Zhang, Zhenru and others},
  title = {The Lessons of Developing Process Reward Models in Mathematical Reasoning},
  year = {2025},
  eprint = {2501.07301},
  archivePrefix = {arXiv}
}

@misc{zhang2025structvrm,
  author = {Zhang, Xiangxiang and others},
  title = {{StructVRM}: Aligning Multimodal Reasoning with Structured and Verifiable Reward Models},
  year = {2025},
  eprint = {2508.05383},
  archivePrefix = {arXiv}
}

@misc{zhao2025genprm,
  author = {Zhao, Jian and others},
  title = {{GenPRM}: Scaling Test-Time Compute of Process Reward Models via Generative Reasoning},
  year = {2025},
  eprint = {2504.00891},
  archivePrefix = {arXiv}
}

@misc{zou2025reasonflux,
  author = {Zou, Jiaru and others},
  title = {{ReasonFlux-PRM}: Trajectory-Aware {PRMs} for Long Chain-of-Thought Reasoning in {LLMs}},
  year = {2025},
  eprint = {2506.18896},
  archivePrefix = {arXiv}
}

@misc{wu2026processthinker,
  author = {Wu, Jingpei and Han, Xiao and Shen, Weixiang and Zhang, Boer and Ding, Zifeng and Tresp, Volker},
  title = {{ProcessThinker}: Enhancing Multi-Modal Large Language Models Reasoning via Rollout-Based Process Reward},
  year = {2026},
  eprint = {2606.11209},
  archivePrefix = {arXiv}
}

@misc{xiong2024llavacritic,
  author = {Xiong, Tianyi and others},
  title = {{LLaVA-Critic}: Learning to Evaluate Multimodal Models},
  year = {2024},
  eprint = {2410.02712},
  archivePrefix = {arXiv}
}

@misc{chen2025rmr1,
  author = {Chen, Xiusi and others},
  title = {{RM-R1}: Reward Modeling as Reasoning},
  year = {2025},
  eprint = {2505.02387},
  archivePrefix = {arXiv}
}

@misc{wang2025unifiedreward,
  author = {Wang, Yibin and others},
  title = {Unified Multimodal Chain-of-Thought Reward Model through Reinforcement Fine-Tuning},
  year = {2025},
  eprint = {2505.03318},
  archivePrefix = {arXiv}
}

@misc{zang2025ixcreward,
  author = {Zang, Yuhang and others},
  title = {{InternLM-XComposer2.5-Reward}: A Simple Yet Effective Multi-Modal Reward Model},
  year = {2025},
  eprint = {2501.12368},
  archivePrefix = {arXiv}
}

@misc{zhang2025mmrlhf,
  author = {Zhang, Yi-Fan and others},
  title = {{MM-RLHF}: The Next Step Forward in Multimodal {LLM} Alignment},
  year = {2025},
  eprint = {2502.10391},
  archivePrefix = {arXiv}
}

@misc{zhang2025r1reward,
  author = {Zhang, Yi-Fan and others},
  title = {{R1-Reward}: Training Multimodal Reward Model through Stable Reinforcement Learning},
  year = {2025},
  eprint = {2505.02835},
  archivePrefix = {arXiv}
}

@misc{wang2026msrl,
  author = {Wang, Chenglong and others},
  title = {{MSRL}: Scaling Generative Multimodal Reward Modeling via Multi-Stage Reinforcement Learning},
  year = {2026},
  eprint = {2603.25108},
  archivePrefix = {arXiv}
}

@misc{li2024vlrewardbench,
  author = {Li, Lei and others},
  title = {{VLRewardBench}: A Challenging Benchmark for Vision-Language Generative Reward Models},
  year = {2024},
  eprint = {2411.17451},
  archivePrefix = {arXiv}
}

@misc{yasunaga2025multimodalrewardbench,
  author = {Yasunaga, Michihiro and Zettlemoyer, Luke and Ghazvininejad, Marjan},
  title = {{Multimodal RewardBench}: Holistic Evaluation of Reward Models for Vision Language Models},
  year = {2025},
  eprint = {2502.14191},
  archivePrefix = {arXiv}
}

@misc{bai2022constitutional,
  author = {Bai, Yuntao and others},
  title = {Constitutional {AI}: Harmlessness from {AI} Feedback},
  year = {2022},
  eprint = {2212.08073},
  archivePrefix = {arXiv}
}

@misc{yu2024criticrm,
  author = {Yu, Yue and others},
  title = {Self-Generated Critiques Boost Reward Modeling for Language Models},
  year = {2024},
  eprint = {2411.16646},
  archivePrefix = {arXiv}
}

@misc{deng2025openvlthinker,
  author = {Deng, Yihe and Bansal, Hritik and Yin, Fan and Peng, Nanyun and Wang, Wei and Chang, Kai-Wei},
  title = {{OpenVLThinker}: An Early Exploration to Complex Vision-Language Reasoning via Iterative Self-Improvement},
  year = {2025},
  eprint = {2503.17352},
  archivePrefix = {arXiv}
}

@misc{ding2025sherlock,
  author = {Ding, Yi and Zhang, Ruqi},
  title = {{Sherlock}: Self-Correcting Reasoning in Vision-Language Models},
  year = {2025},
  eprint = {2505.22651},
  archivePrefix = {arXiv}
}

@misc{liu2025noisyrollout,
  author = {Liu, Xiangyan and others},
  title = {{NoisyRollout}: Reinforcing Visual Reasoning with Data Augmentation},
  year = {2025},
  eprint = {2504.13055},
  archivePrefix = {arXiv}
}

@misc{liu2025spark,
  author = {Liu, Ziyu and others},
  title = {{SPARK}: Synergistic Policy and Reward Co-Evolving Framework},
  year = {2025},
  eprint = {2509.22624},
  archivePrefix = {arXiv}
}

@misc{ma2025s2r,
  author = {Ma, Ruotian and others},
  title = {{S2R}: Teaching {LLMs} to Self-Verify and Self-Correct via Reinforcement Learning},
  year = {2025},
  eprint = {2502.12853},
  archivePrefix = {arXiv}
}

@misc{wan2025srpo,
  author = {Wan, Zhongwei and others},
  title = {{SRPO}: Enhancing Multimodal {LLM} Reasoning via Reflection-Aware Reinforcement Learning},
  year = {2025},
  eprint = {2506.01713},
  archivePrefix = {arXiv}
}

@misc{wang2025sotawithless,
  author = {Wang, Xiyao and others},
  title = {{SoTA} with Less: {MCTS}-Guided Sample Selection for Data-Efficient Visual Reasoning Self-Improvement},
  year = {2025},
  eprint = {2504.07934},
  archivePrefix = {arXiv}
}

@misc{wang2025vlrethinker,
  author = {Wang, Haozhe and Qu, Chao and Huang, Zuming and Chu, Wei and Lin, Fangzhen and Chen, Wenhu},
  title = {{VL-Rethinker}: Incentivizing Self-Reflection of Vision-Language Models with Reinforcement Learning},
  year = {2025},
  eprint = {2504.08837},
  archivePrefix = {arXiv}
}

@misc{lu2026tgrl,
  author = {Lu, Jinda and others},
  title = {Beyond Where to Look: Trajectory-Guided Reinforcement Learning for Multimodal {RLVR}},
  year = {2026},
  eprint = {2603.26126},
  archivePrefix = {arXiv}
}

@misc{tang2026vgrreflection,
  author = {Tang, Liyan and Yin, Fangcong and Durrett, Greg},
  title = {Visually Grounded Self-Reflection for Vision-Language Models via Reinforcement Learning},
  year = {2026},
  eprint = {2607.02490},
  archivePrefix = {arXiv}
}

@misc{lu2023mathvista,
  author = {Lu, Pan and others},
  title = {{MathVista}: Evaluating Mathematical Reasoning of Foundation Models in Visual Contexts},
  year = {2023},
  eprint = {2310.02255},
  archivePrefix = {arXiv}
}

@misc{yue2023mmmu,
  author = {Yue, Xiang and others},
  title = {{MMMU}: A Massive Multi-Discipline Multimodal Understanding and Reasoning Benchmark for Expert {AGI}},
  year = {2023},
  eprint = {2311.16502},
  archivePrefix = {arXiv}
}

@inproceedings{wang2024mathvision,
  author = {Wang, Ke and others},
  title = {Measuring Multimodal Mathematical Reasoning with {MATH-Vision} Dataset},
  booktitle = {Advances in Neural Information Processing Systems, Datasets and Benchmarks Track},
  year = {2024}
}

@misc{yue2024mmmupro,
  author = {Yue, Xiang and others},
  title = {{MMMU-Pro}: A More Robust Multi-Discipline Multimodal Understanding Benchmark},
  year = {2024},
  eprint = {2409.02813},
  archivePrefix = {arXiv}
}

@misc{zhang2024mathverse,
  author = {Zhang, Renrui and others},
  title = {{MathVerse}: Does Your Multi-Modal {LLM} Truly See the Diagrams in Visual Math Problems?},
  year = {2024},
  eprint = {2403.14624},
  archivePrefix = {arXiv}
}

@article{li2026unim,
  title = {{UniM}: A Unified Any-to-Any Interleaved Multimodal Benchmark},
  author = {Li, Yanlin and Guo, Minghui and Zhang, Kaiwen and Zhang, Shize and Zhao, Yiran and Li, Haodong and Zhou, Congyue and Zheng, Weijie and Yan, Yushen and Wu, Shengqiong and others},
  journal = {arXiv preprint arXiv:2603.05075},
  year = {2026}
}

\end{document}